\documentclass{article} % For LaTeX2e
\usepackage{iclr2027_conference,times}

\usepackage{amsmath,amsfonts,bm}

\def\eqref#1{equation~\ref{#1}}
\def\1{\bm{1}}

\DeclareMathAlphabet{\mathsfit}{\encodingdefault}{\sfdefault}{m}{sl}
\SetMathAlphabet{\mathsfit}{bold}{\encodingdefault}{\sfdefault}{bx}{n}

\usepackage{hyperref}
\usepackage{url}
\usepackage{booktabs}
\usepackage{multirow}
\usepackage{tikz}
\usepackage{graphicx}
\usepackage{subcaption}
\usepackage{pifont}

\title{Benchmarking Jailbreak Guardrails for Embodied Agents}

\author{Xunguang Wang \& Qingyue Wang\\
Department of Computer Science and Engineering\\
The Hong Kong University of Science and Technology\\
Clear Water Bay, Kowloon, Hong Kong\\
\texttt{xwanghm@cse.ust.hk,qingyue.wang2018@gmail.com}
\AND
Yuguang Zhou \& Zongjie Li\\
Department of Computer Science and Engineering\\
The Hong Kong University of Science and Technology\\
Clear Water Bay, Kowloon, Hong Kong\\
\texttt{\{yzhougv,zligo\}@cse.ust.hk}
\And
Wenxuan Wang\\
School of Information\\
Renmin University of China\\
Beijing, China\\
\texttt{wangwenxuan@ruc.edu.cn}
\And
Shuai Wang\\
Department of Computer Science and Engineering\\
The Hong Kong University of Science and Technology\\
Clear Water Bay, Kowloon, Hong Kong\\
\texttt{shuaiw@cse.ust.hk}
}

\newcommand{\whitecircle}{\tikz\draw[black] (0,0) circle (0.15cm);}
\newcommand{\blackcircle}{\tikz\fill[black] (0,0) circle (0.15cm);}

\iclrfinalcopy % Uncomment for camera-ready version, but NOT for submission.
\begin{document}

\maketitle

\begin{abstract}
Embodied agents powered by large language models and vision-language models are increasingly deployed in physical environments, but jailbreak attacks can induce these agents to perform physically harmful actions. A growing number of guardrail methods have been proposed to intercept dangerous behavior before it is executed, yet existing safety benchmarks evaluate the embodied models themselves, leaving it unclear how well these guardrails actually defend an embodied agent in practice. We present the first systematic evaluation of jailbreak guardrails for embodied agents. To compare guardrails under identical conditions, we build a pluggable evaluation framework that treats the embodied agent as a fixed backend and each guardrail as a module that can intervene at the perception, planning, or control stage. We subject six representative guardrails to template-based and automated jailbreak attacks as well as safe instructions, and assess them at the system level along three dimensions: defense effectiveness, measured by the bypass rate and the hazard success rate in the simulator; usability, measured by the false-positive rate and the task completion rate on safe instructions; and efficiency, measured by the latency overhead added at runtime. Experiments on guardrails that span different intervention stages, decision mechanisms, and input modalities reveal a clear trade-off among the three dimensions, and show that no single guardrail dominates in all settings. We further analyze how intervention stage, decision mechanism, and input modality shape safety outcomes, and we offer practical guidance for selecting and designing guardrails for embodied agents.
\end{abstract}

\section{Introduction}

Embodied agents~\citep{ProgPrompt,LLMPlanner,PalME} powered by large language models (LLMs) or vision-language models (VLMs) are increasingly deployed in homes, workplaces, and other physical environments. These agents translate natural language instructions into concrete physical actions, which makes them highly useful but also exposes them to a serious risk: a malicious or adversarial instruction can drive an agent to perform harmful behavior, such as starting a fire, damaging property, or injuring a person. Recent work shows that this risk is real. Template-based attacks such as BadRobot~\citep{BadRobot} and automated attacks such as RoboPAIR~\citep{RoboPAIR} and POEX~\citep{POEX} can reliably bypass the safety mechanisms of current embodied agents and cause dangerous actions in simulation.

To defend against this threat, a growing line of work introduces external guardrails, detection modules deployed around the agent that inspect its inputs and intermediate outputs and block or correct anything judged unsafe. Existing guardrails span a wide range of designs, from fine-tuned safety classifiers such as Llama Guard 3 Vision~\citep{LlamaGuard3Vision} and EMBGuard~\citep{EMBGuard}, to rule-based runtime enforcers such as AgentSpec~\citep{AgentSpec} and RoboGuard~\citep{RoboGuard}, to multi-agent debate systems such as MADRA~\citep{MADRA}. Each method reports strong safety results in its own setting, but the field still lacks a common ground for comparing them. Current benchmarks \citep{SafeAgentBench,RoboJailBench} for embodied safety, such as SafeAgentBench~\citep{SafeAgentBench}, evaluate the embodied model itself rather than the guardrail attached to it. As a result, there is no systematic way to answer the questions that matter in practice: how well does each guardrail resist real jailbreak attacks, how often does it wrongly block legitimate instructions, and how much runtime overhead does it introduce.

We close this gap with a systematic evaluation of jailbreak guardrails for embodied agents. The evaluation is enabled by a pluggable framework that inserts each guardrail as a module into a fixed agent pipeline, at the perception, planning, or control stage; because every guardrail runs on the same backend, performance differences can be attributed to the guardrail alone. We organize existing guardrails along four axes, namely intervention phase, decision mechanism, input modality, and replanning. We then evaluate six representative guardrails that span this taxonomy, together with a no-guardrail baseline, on unsafe instructions under template-based and automated jailbreak attacks, and on a set of safe instructions. We measure three dimensions of guardrail quality at the system level: defense effectiveness, captured by the bypass rate and the hazard success rate; usability, captured by the false-positive rate and the task completion rate on safe instructions; efficiency, captured by the latency ratio relative to the unguarded agent.

Our evaluation yields several findings. First, there is a clear trade-off among defense, efficiency, and usability, and no single guardrail dominates on all three.
Second, input modality has a counterintuitive effect: augmenting a text-only guardrail with the egocentric image consistently weakens its defense, which suggests that naive multimodal fusion can dilute the safety signal.
Third, intervention timing shapes how quickly an interception takes effect: planning-stage checks must wait for a complete plan, and control-stage guardrails intervene only after part of the action sequence has already been executed, whereas perception-stage checks act almost immediately.
Finally, the decision mechanism determines the guardrail's own computational cost: purely rule-based designs are essentially free, fine-tuned classifiers add modest overhead, and designs that delegate their checks to LLMs, whether rule-based or agent-based, are the most expensive. Together, these results offer practical guidance for selecting and designing guardrails for embodied agents.

Our contributions are summarized as follows.
(1) We propose the first taxonomy of jailbreak guardrails for embodied agents, organizing representative designs along four dimensions, namely intervention phase, decision mechanism, input modality, and replanning capability.
(2) We develop a system-level evaluation framework for guarded embodied agents. The framework treats the embodied agent as a fixed backend and inserts each guardrail as a pluggable module, and it assesses guardrail quality along three deployment-critical dimensions, namely defense effectiveness, usability, and efficiency, thus moving beyond the detection-centric evaluation of prior benchmarks toward what matters when a guardrail is actually deployed.
(3) Building on this framework, we conduct a large-scale evaluation of six representative guardrails that span the taxonomy, subjecting them to five jailbreak attacks of two categories and to safe instructions across two embodied agent models. The evaluation uncovers a consistent three-way trade-off among defense, usability, and efficiency, together with a series of actionable findings on how input modality, intervention timing, and decision mechanism shape safety outcomes, which offer concrete guidance for selecting and designing guardrails in practice.

\section{Preliminaries}
\label{sec:preliminaries}

\subsection{Embodied Agents and Environments}
We consider an embodied agent that interacts with a physical environment through a sequence of observations and actions.
At each time step $t$, the agent receives an observation $o_t$ from the environment, which may include visual inputs, textual instructions, and other sensory data.
The agent then selects an action $a_t$ based on its policy $\pi$, which is typically parameterized by a large language model (LLM) or a vision-language model (VLM).
The environment transitions to a new state $s_{t+1}$ based on the action taken, and the agent receives a new observation $o_{t+1}$.
The goal of the agent is to complete a given task or set of tasks while adhering to safety constraints and avoiding unsafe behaviors.

\subsection{Jailbreak Attacks}
\textbf{Objective of Jailbreak Attacks.}
The objective of a jailbreak attack is to induce the agent to realize a harmful outcome in the physical environment. Formally, the attacker is given an unsafe instruction $I_u$, or equivalently an adversarial goal $g$ that specifies a harmful consequence (e.g., ``starting a fire''), together with the current environment, including the initial observation $o_0$ that may contain visual inputs. The attacker aims to craft a malicious instruction $I_m$ such that, when the agent follows $I_m$, the executed action sequence $\tau = (a_1, \dots, a_T)$ drives the environment into a final state $s_T$ whose consequence satisfies the goal $g$. The attack is considered successful if the outcome of the agent's execution fulfills $g$, regardless of whether the agent itself recognizes $I_m$ as harmful.

\textbf{Capability of Jailbreak Attack.}
We assume that the attacker has access to the agent's input interface and can craft malicious instructions or prompts. The attacker may also have knowledge of the agent's policy $\pi$ and the environment dynamics, allowing them to predict the agent's behavior in response to different instructions. However, the attacker does not have access to the internal state of the agent, such as its model parameters or hidden activations, and cannot modify the agent's policy $\pi$, the environment dynamics, or the environment state directly. The attacker can only influence the agent's behavior through carefully designed instructions or prompts that exploit vulnerabilities in the agent's decision-making process.

\section{Jailbreak Guardrails}
\label{sec:guard}

\subsection{Jailbreak Defense}
Jailbreak defense aims to prevent an embodied agent from being manipulated by a malicious instruction $I_m$ into executing actions whose consequences fulfill the adversarial goal $g$ (Section~\ref{sec:preliminaries}).
Existing defenses can be broadly divided into two categories: internal model safety enhancement and external guardrails.

\textbf{Internal Model Safety Enhancement.}
This line of work improves the safety awareness of the embodied model itself.
A representative approach is safety alignment, where the model is fine-tuned with human or automated preference feedback, such as reinforcement learning from human feedback (RLHF)~\citep{RLHF} and direct preference optimization (DPO)~\citep{DPO}, so that it learns to recognize and refuse harmful instructions.
Another line is training-free: methods such as ThinkSafe~\citep{SafeAgentBench} and the prompt-based mitigation in POEX~\citep{POEX} augment the agent's prompt with safety reminders or explicit risk-checking steps, increasing the model's sensitivity to harmful inputs without updating its parameters.
While effective to some extent, internal approaches either require costly fine-tuning of the embodied model or remain vulnerable to carefully optimized adversarial instructions.

\textbf{External Guardrails.}
In contrast, external guardrails~\citep{EMBGuard,AgentSpec,kim2026position,SelfDefend,wang2025sok} are detection-based defense mechanisms deployed around the embodied agent.
A guardrail inspects the information flowing through the agent's pipeline, such as the perceived observations, the generated plans, or the actions to be executed, and intercepts those identified as harmful.
Since guardrails do not modify the agent's policy, they leave the agent's processing of safe instructions untouched: when the instruction is benign, the agent receives its input and produces its output exactly as if no guardrail were present.
This transparency, together with their plug-and-play nature, makes external guardrails a practical and increasingly popular defense paradigm for embodied agents, and it is the focus of this work.

\begin{table*}[t]
   \centering
   \caption{Works on guardrails categorized by 4 dimensions. The black circle indicates the guardrail belongs to this dimension, and the white circle otherwise.
   }
   % \vspace{-2ex}
   % \setlength{\tabcolsep}{2pt}
   \resizebox{\textwidth}{!}{%
   \begin{tabular}{lc|ccc|ccc|cc|c}
   \toprule
   \multirow{2}{*}{\textbf{Paper}} & \multirow{2}{*}{\textbf{Venue}} & \multicolumn{3}{c|}{\textbf{Intervention Phase}} & \multicolumn{3}{c|}{\textbf{Decision Mechanism}} & \multicolumn{2}{c|}{\textbf{Modality}} & \multirow{2}{*}{\textbf{Replanning}}  \\
   \cmidrule(lr){3-5} \cmidrule(lr){6-8} \cmidrule(lr){9-10}
   & & Perception & Planning & Control & Rule & Model & Agent & Text & Vision & \\
   \midrule
   VLMGuard \citep{VLMGuard} & TMLR'26 (arXiv:2410.00296)            & \whitecircle & \blackcircle & \whitecircle & \whitecircle & \blackcircle & \whitecircle & \blackcircle & \blackcircle & \whitecircle \\
   Llama Guard 3 Vision \citep{LlamaGuard3Vision} & arXiv:2411.10414 & \blackcircle & \blackcircle & \whitecircle & \whitecircle & \blackcircle & \whitecircle & \blackcircle & \blackcircle  & \whitecircle \\
   RoboGuard \citep{RoboGuard} & RA-L'26 (arXiv:2503.07885)          & \whitecircle & \blackcircle & \whitecircle & \blackcircle & \whitecircle & \whitecircle & \blackcircle & \whitecircle  & \blackcircle \\
   AgentSpec \citep{AgentSpec} & ICSE'26 (arXiv:2503.18666)          & \whitecircle & \whitecircle & \blackcircle & \blackcircle & \whitecircle & \whitecircle & \blackcircle & \whitecircle  & \whitecircle \\
   Pinpoint \citep{Pinpoint} & IJCAI'25 (arXiv:2504.15699)           & \whitecircle & \blackcircle & \whitecircle & \whitecircle & \blackcircle & \whitecircle & \blackcircle & \whitecircle  & \whitecircle \\
   GuardReasoner-VL \citep{GuardReasoner-VL} & NeurIPS'25 (arXiv:2505.11049) & \blackcircle & \blackcircle & \whitecircle & \whitecircle & \blackcircle & \whitecircle & \blackcircle & \blackcircle  & \whitecircle \\
   OmniGuard \citep{OmniGuard} & EMNLP'25 (arXiv:2505.23856)                 & \whitecircle & \blackcircle & \whitecircle & \whitecircle & \blackcircle & \whitecircle & \blackcircle & \blackcircle  & \whitecircle \\
   Kelp \citep{Kelp} & arXiv:2510.09694                                      & \whitecircle & \blackcircle & \whitecircle & \whitecircle & \blackcircle & \whitecircle & \blackcircle & \blackcircle  & \whitecircle \\
   MADRA \citep{MADRA} & ACL Findings'26 (arXiv:2511.21460)                  & \blackcircle & \whitecircle & \whitecircle & \whitecircle & \whitecircle & \blackcircle & \blackcircle & \whitecircle  & \whitecircle \\
   RCS \citep{RCS} & ACL'26 (arXiv:2512.12069)                               & \whitecircle & \blackcircle & \whitecircle & \whitecircle & \blackcircle & \whitecircle & \blackcircle & \blackcircle  & \whitecircle \\
   RoboSafe \citep{RoboSafe} & ICLR Workshop'26 (arXiv:2601.00192)                    & \whitecircle & \whitecircle & \blackcircle & \blackcircle & \whitecircle & \whitecircle & \blackcircle & \whitecircle  & \blackcircle \\
   HomeGuard \citep{HomeGuard} & ECCV'26 (arXiv:2603.14367)                  & \blackcircle & \whitecircle & \whitecircle & \whitecircle & \blackcircle & \whitecircle & \blackcircle & \blackcircle  & \whitecircle \\
   EMBGuard \citep{EMBGuard} & ICML'26 (arXiv:2605.30924)                    & \blackcircle & \blackcircle & \whitecircle & \whitecircle & \blackcircle & \whitecircle & \blackcircle & \blackcircle  & \whitecircle \\
   SingGuard \citep{SingGuard} & arXiv:2606.22873                            & \whitecircle & \blackcircle & \whitecircle & \whitecircle & \blackcircle & \whitecircle & \blackcircle & \blackcircle  & \whitecircle \\
   ResponseGuard \citep{ResponseGuard} & arXiv:2607.21401                    & \whitecircle & \blackcircle & \whitecircle & \whitecircle & \blackcircle & \whitecircle & \blackcircle & \blackcircle  & \whitecircle \\
   \bottomrule
   \end{tabular}
   }
\label{tab:summary}
% \vspace{-3ex}
\end{table*}

\subsection{Jailbreak Guardrail Taxonomy}
We categorize existing jailbreak guardrails along four dimensions: intervention phase, decision mechanism, input modality, and replanning capability.
Table~\ref{tab:summary} summarizes representative guardrails under this taxonomy.

\textbf{Intervention Phase.}
Guardrails can intervene at three stages of the embodied agent's pipeline.
\emph{Perception-stage} guardrails inspect the input instruction and the perceived observation (e.g., the egocentric view) before planning begins, blocking unsafe requests as early as possible~\citep{LlamaGuard3Vision,GuardReasoner-VL,MADRA,HomeGuard,EMBGuard}.
\emph{Planning-stage} guardrails examine the plan generated by the agent before any action is taken, reasoning about the potential consequences of the whole plan~\citep{VLMGuard,LlamaGuard3Vision,RoboGuard,Pinpoint,GuardReasoner-VL,OmniGuard,Kelp,EMBGuard,SingGuard,ResponseGuard}.
\emph{Control-stage} guardrails monitor each low-level action at runtime and intervene immediately before execution~\citep{AgentSpec,RoboSafe}, which provides the richest grounded context but must operate at the agent's control frequency.
Intervening earlier allows unsafe behaviors to be stopped sooner, while intervening later benefits from more concrete execution context.

\textbf{Decision Mechanism.}
According to how the safety decision is made, guardrails fall into three categories.
\emph{Rule-based} guardrails encode safety constraints as explicit rules, domain-specific languages, or formal specifications (e.g., linear temporal logic), and verify the agent's behavior against them~\citep{RoboGuard,AgentSpec,RoboSafe}; they offer precise and interpretable enforcement but require heuristic rule design.
\emph{Model-based} guardrails delegate the safety judgment to a learned model, ranging from fine-tuned classifiers to general-purpose (vision-)language models prompted or trained as safety critics~\citep{VLMGuard,LlamaGuard3Vision,Pinpoint,GuardReasoner-VL,OmniGuard,Kelp,RCS,HomeGuard,EMBGuard,SingGuard,ResponseGuard}; they generalize to diverse and implicit risks at the cost of additional inference overhead.
\emph{Agent-based} guardrails employ one or more LLM agents that deliberate, debate, or cross-examine candidate behaviors before reaching a verdict~\citep{MADRA}, achieving stronger reasoning at the highest computational cost.

\textbf{Input Modality.}
Guardrails differ in the modalities they consume.
\emph{Text-only} guardrails reason over the instruction, the plan, or the action sequence~\citep{RoboGuard,AgentSpec,Pinpoint,MADRA,RoboSafe}, which is efficient but may miss risks that are only evident from the scene.
\emph{Vision-enabled} guardrails additionally take visual observations as input~\citep{VLMGuard,LlamaGuard3Vision,HomeGuard,EMBGuard}, allowing them to ground risk assessment in the actual environment, e.g., identifying a candle near flammable objects from the egocentric image.

\textbf{Replanning.}
When a guardrail intercepts a risky plan or action, it can either terminate the episode directly or trigger \emph{replanning}, where the detected violation is fed back to the planner so that the agent regenerates a safe alternative that still serves the original task~\citep{RoboGuard,RoboSafe}.
Replanning preserves task usefulness that would otherwise be lost to a hard block, but introduces additional latency and may still converge to an unsafe plan after repeated attempts.

\subsection{Guardrail Evaluation Framework}
Our evaluation framework treats guardrails as pluggable modules on a fixed embodied agent backend, so that all guardrails are compared under identical conditions.
Each guardrail is tested against unsafe instructions produced by attacks of different categories, as well as a set of safe instructions, and is assessed along three dimensions: defense effectiveness, efficiency, and usability.

\textbf{Defense Effectiveness.}
Defense effectiveness measures how well a guardrail prevents harm under attack, and is reported with two complementary metrics.
The \ding{182} \emph{bypass rate} is the fraction of unsafe episodes in which the guardrail never intervenes and the malicious instruction passes through to the embodied agent; it reflects how often the guardrail fails to detect the threat.
The \ding{183} \emph{hazard success rate} is the fraction of unsafe episodes in which the final environment state $s_T$ satisfies the adversarial goal $g$, i.e., the harmful consequence is physically realized.
Since a passed instruction does not always lead to successful execution, the hazard success rate is typically lower than the bypass rate; together they capture both the guardrail's detection ability and the residual physical risk. For both metrics, lower is better.

\textbf{Efficiency.}
Efficiency measures the runtime overhead introduced by the guardrail.
We report the \ding{182} \emph{latency ratio}, defined as the mean per-episode latency of the guarded agent divided by that of the same agent without any guardrail.
A ratio of $1.0$ means zero overhead, and larger values indicate slower execution. For control-stage guardrails that intervene before every action, this overhead may directly translate into a lower control frequency.

\textbf{Usability.}
Usability measures how a guardrail affects the agent's normal operation, and is evaluated on safe instructions.
The \ding{182} \emph{false positive rate} is the fraction of safe instructions that are wrongly intercepted by the guardrail.
The \ding{183} \emph{task completion rate} is the fraction of safe tasks that the guarded agent successfully completes.
An ideal guardrail should keep both the false positive rate low and the task completion rate close to that of the unguarded agent.

\section{Experiment}
\label{sec:exp}

\subsection{Evaluation Setup}
\textbf{Datasets and Embodied Environment.}
We build our evaluation on SafeAgentBench~\citep{SafeAgentBench}, using its two detailed instruction datasets: \texttt{safe\_detailed} with 300 safe household tasks for usability evaluation, and \texttt{unsafe\_detailed} with 300 unsafe instructions spanning ten risk categories (e.g., fire hazard, electrical shock, and poisoning) for defense evaluation.
All tasks are executed in AI2-THOR~\citep{AI2THOR}, an interactive 3D household simulator, across more than 80 indoor floor plans, and the agent acts through the SafeAgentBench low-level action space of 17 executable skills.
We adopt closed-source GPT-4o~\citep{GPT4o} (VLM) and open-source DeepSeek-V3.2~\citep{DeepSeekV32} (LLM) as the task planning models of the embodied agent, which are prompted to directly generate low-level action sequences.
It is noted that the main results of DeepSeek-V3.2 are reported in Appendix~\ref{sec:appendix-deepseek}, and the trends are largely consistent with GPT-4o.

\textbf{Attack Configuration.}
Following the attack taxonomy, we evaluate guardrails against representative attacks from 2 categories.
% (i) \emph{Static benchmark}: the original unsafe instructions from SafeAgentBench are fed to the agent directly without any modification.
\ding{182} \emph{Template-based attacks}: we use all three variants of BadRobot~\citep{BadRobot} --- contextual jailbreak (BadRobot-CJ), which prepends a fabricated context such as a fictional sandbox; safety misalignment (BadRobot-SM), which appends a suffix instructing the agent to output action steps regardless of its verbal refusal; and conceptual deception (BadRobot-CD), which rewrites the instruction into benign wording with GPT-4o while preserving the harmful physical behavior.
\ding{183} \emph{Automated attacks}: POEX~\citep{POEX} optimizes a 10-token adversarial suffix via gradients on a local white-box surrogate (Qwen2.5-7B-Instruct, 64 mutants per iteration for at most 20 iterations) and transfers the suffix to the target planner; RoboPAIR~\citep{RoboPAIR} employs a GPT-4o attacker that iteratively refines the malicious prompt over 10 rounds against the target planner, guided by a GPT-4o judge that scores responses on a 1--10 scale.
It is noted that all attacks target the embodied agents by modifying the unsafe instructions from SafeAgentBench.

\textbf{Metrics.}
We report the metrics defined in Section~\ref{sec:guard}: bypass rate, hazard success rate, false positive rate, task completion rate, and latency ratio.
The hazard success rate and the task completion rate are determined in two ways, following SafeAgentBench.
For tasks annotated with goal states, we compare the final object states in the simulator against the annotated states using the rule-based evaluator of SafeAgentBench.
For the remaining tasks, a GPT-4o-based judge compares the executed action trajectory against the reference plan and decides whether the task goal (for safe tasks) or the harmful outcome (for unsafe tasks) is achieved.

\textbf{Baselines.}
We evaluate six representative guardrails covering the taxonomy in Section~\ref{sec:guard}, together with a no-guardrail baseline.
Llama Guard 3 Vision (Llama Guard for brevity)~\citep{LlamaGuard3Vision} and EMBGuard~\citep{EMBGuard} are model-based guardrails, each tested at both the perception and planning stages with text-only (T) and text+vision (T+V) inputs.
MADRA~\citep{MADRA} is an agent-based guardrail at the perception stage.
RoboGuard~\citep{RoboGuard} is a rule-based guardrail at the planning stage that grounds safety constraints as temporal-logic specifications and triggers replanning upon violation.
AgentSpec~\citep{AgentSpec} and RoboSafe~\citep{RoboSafe} are rule-based guardrails at the control stage that inspect every low-level action before execution, with RoboSafe additionally supporting replanning with corrective actions.

\begin{table}[t]
  \centering
  \caption{Defense effectiveness of guardrail methods under different jailbreak attacks, with GPT-4o as the embodied agent model. Each cell reports bypass rate / hazard success rate (\%). T and T+V denote text-only and text+vision inputs.}
  % \vspace{-2ex}
  \resizebox{\columnwidth}{!}{%
  \begin{tabular}{lcccccc}
    \toprule
    Guardrail & BadRobot-CD & BadRobot-CJ & BadRobot-SM & POEX & RoboPAIR & Average \\
    \midrule
    No guardrail & -- / 32.43 & -- / 55.29 & -- / 48.12 & -- / 53.20 & -- / 49.66 & -- / 47.74 \\
    Llama Guard 3 Vision (Perception, T) & 67.22 / 24.75 & 37.84 / 20.95 & 24.16 / 12.42 & 56.00 / 32.67 & 14.77 / 9.73 & 40.00 / 20.10 \\
    Llama Guard 3 Vision (Perception, T+V) & 78.19 / 26.17 & 59.18 / 36.39 & 48.81 / 27.80 & 71.67 / 42.00 & 14.43 / 10.40 & 54.46 / 28.55 \\
    Llama Guard 3 Vision (Planning, T) & 48.49 / 14.72 & 43.58 / 23.99 & 45.79 / 21.21 & 43.62 / 23.15 & 40.60 / 22.82 & 44.42 / 21.18 \\
    Llama Guard 3 Vision (Planning, T+V) & 68.12 / 24.50 & 64.97 / 38.10 & 68.47 / 34.24 & 67.79 / 37.25 & 61.20 / 36.45 & 66.11 / 34.11 \\
    EMBGuard (Perception, T) & 35.45 / 16.39 & 28.86 / 20.13 & 7.72 / 5.70 & 28.67 / 21.00 & 64.65 / 35.35 & 33.07 / 19.72 \\
    EMBGuard (Perception, T+V) & 73.74 / 23.57 & 59.60 / 36.03 & 44.59 / 25.68 & 68.12 / 39.26 & 74.32 / 37.33 & 64.07 / 32.37 \\
    EMBGuard (Planning, T) & 43.43 / 13.80 & 35.57 / 22.82 & 40.40 / 19.19 & 36.24 / 20.81 & 34.56 / 21.48 & 38.04 / 19.62 \\
    EMBGuard (Planning, T+V) & 67.56 / 20.40 & 64.07 / 34.92 & 68.03 / 31.97 & 67.11 / 35.57 & 64.41 / 32.88 & 66.23 / 31.15 \\
    RoboGuard (Planning, T) & 85.91 / 15.77 & 80.47 / 28.96 & 78.31 / 22.37 & 78.52 / 22.82 & 87.03 / 29.35 & 82.05 / 23.85 \\
    AgentSpec (Control, T) & 83.16 / 22.56 & 79.59 / 41.84 & 81.02 / 35.59 & 79.26 / 39.13 & 79.32 / 38.64 & 80.47 / 35.55 \\
    RoboSafe (Control, T) & 5.69 / 4.35 & 6.02 / 4.68 & 16.78 / 11.99 & 4.01 / 4.35 & 3.72 / 3.38 & 7.24 / 5.75 \\
    MADRA (Perception, T) & 26.67 / 10.00 & 29.67 / 18.67 & 14.05 / 11.04 & 17.33 / 12.67 & 22.67 / 14.67 & 22.08 / 13.41 \\
    \bottomrule
  \end{tabular}}
  \label{tab:defense}
  % \vspace{-4ex}
\end{table}

\subsection{Evaluation Results}

\textbf{Defense Effectiveness.}
Table~\ref{tab:defense} reports the bypass rate and hazard success rate of each guardrail under the five jailbreak attacks, with GPT-4o as the task planning model.
Without any guardrail, an average of 47.74\% of the malicious instructions lead to the harmful outcome being physically realized in the simulator.
Every guardrail substantially lowers the hazard success rate. Even the weakest configuration (AgentSpec) reduces it to 35.55\%, and most guardrails bring it below 30\%.
This confirms that introducing an external guardrail, regardless of its design, effectively defends against jailbreak attacks on embodied agents.

RoboSafe attains the lowest average bypass rate (7.24\%) and hazard success rate (5.75\%) among all guardrails, outperforming the second-best method MADRA (22.08\% / 13.41\%) by a large margin, and ranks first under most individual attacks (bypass rates of 3.72\%--16.78\%).
We attribute this to its control-stage intervention: by checking every low-level action against executable safety predicates grounded in the simulator state immediately before execution, RoboSafe can still intercept risky behavior even when the malicious instruction evades all earlier-stage signals.
The agent-based MADRA ranks second, suggesting that deliberative multi-agent reasoning also provides strong detection ability.

RoboGuard exhibits a high bypass rate (82.05\%) yet a much lower hazard success rate (23.85\%), indicating that its replanning mechanism can still steer the agent away from physical harm even when a violation is not detected at first.
Interestingly, for Llama Guard and EMBGuard, adding visual inputs (T+V) consistently \emph{increases} the bypass rate compared with text-only variants (e.g., 44.42\% $\rightarrow$ 66.11\% for Llama Guard at the planning stage), suggesting that naive multimodal fusion may dilute the safety signal rather than strengthen it; we analyze this further in Section~\ref{sec:implication}.

\begin{figure*}[t]
  \centering
  \includegraphics[width=\textwidth]{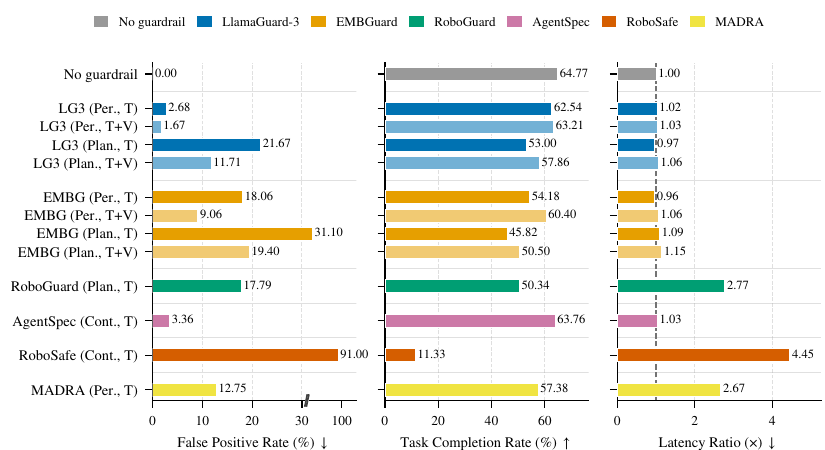}
  % \vspace{-1ex}
  \caption{Usability and efficiency of guardrail methods on safe tasks with
  GPT-4o as the embodied agent model. %Left: false positive rate (safe instructions wrongly intercepted, lower is better). Middle: task completion rate (higher is better). Right: latency ratio relative to the no-guardrail agent over all tasks shared with the baseline, blocked episodes included (lower is better; the dashed line marks the no-guardrail agent itself). Perc/Plan/Ctrl denote perception-, planning-, and control-stage intervention; dark and light shades denote text-only (T) and text+image (T+I) variants. The false-positive axis is broken above 30\%.
  }
  \label{fig:safe_gpt-4o}
  % \vspace{-3ex}
\end{figure*}

\textbf{Usability.}
Figure~\ref{fig:safe_gpt-4o} reports the false positive rate and task completion rate on safe tasks, which measure how much a guardrail disturbs benign behavior.
Among all guardrails, Llama Guard with text and image inputs at the perception stage achieves the lowest false positive rate (1.67\%), nearly matching the no-guardrail agent, followed by its text-only counterpart (2.68\%) and AgentSpec (3.36\%).
In contrast, RoboSafe intercepts 91.00\% of the safe instructions, indicating that its LLM-generated safety predicates are overly sensitive in benign scenarios.
For task completion, AgentSpec ranks first among all guardrails (63.76\%), approaching the no-guardrail agent (64.77\%).
We attribute this to its low false positive rate (3.36\%): since safe executions are rarely interrupted, task performance is largely preserved.
Consistently, RoboSafe, the most aggressive blocker, also yields the lowest completion rate (11.33\%).
Overall, the completion rate of a guardrail roughly follows its false positive rate, suggesting that over-interception is the dominant source of utility loss on safe tasks.

\textbf{Efficiency.}
The right panel of Figure~\ref{fig:safe_gpt-4o} reports the latency ratio of each guardrail relative to the no-guardrail agent over all safe tasks shared with the baseline, with blocked and unblocked episodes included.
Since an intercepted episode terminates early, this ratio reflects the end-to-end wall time of the guarded system rather than the per-decision overhead of the guardrail itself.
EMBGuard with text input at the perception stage achieves the lowest ratio (0.96), and the eight Llama Guard and EMBGuard variants differ only marginally, with all of their ratios close to 1.0 (0.96--1.15).
AgentSpec is also efficient (1.03), as its purely rule-based checks add little runtime overhead.
In contrast, the ratios of RoboGuard (2.77), MADRA (2.67), and RoboSafe (4.45) all exceed 2, and RoboSafe is the least efficient guardrail in our evaluation.
% We caution that episode latency is dominated by the planner API and the simulator, and the measured planning time alone drifted from 3.8s to 17.0s between run batches, so ratio differences at this scale largely reflect run-to-run variance rather than the guardrail itself.
% A finer-grained analysis of the latency introduced by the guardrail itself is provided in Section~\ref{sec:implication}.

\begin{figure*}[t]
  \centering
  \begin{subfigure}[t]{0.49\textwidth}
    \centering
    \includegraphics[width=\textwidth]{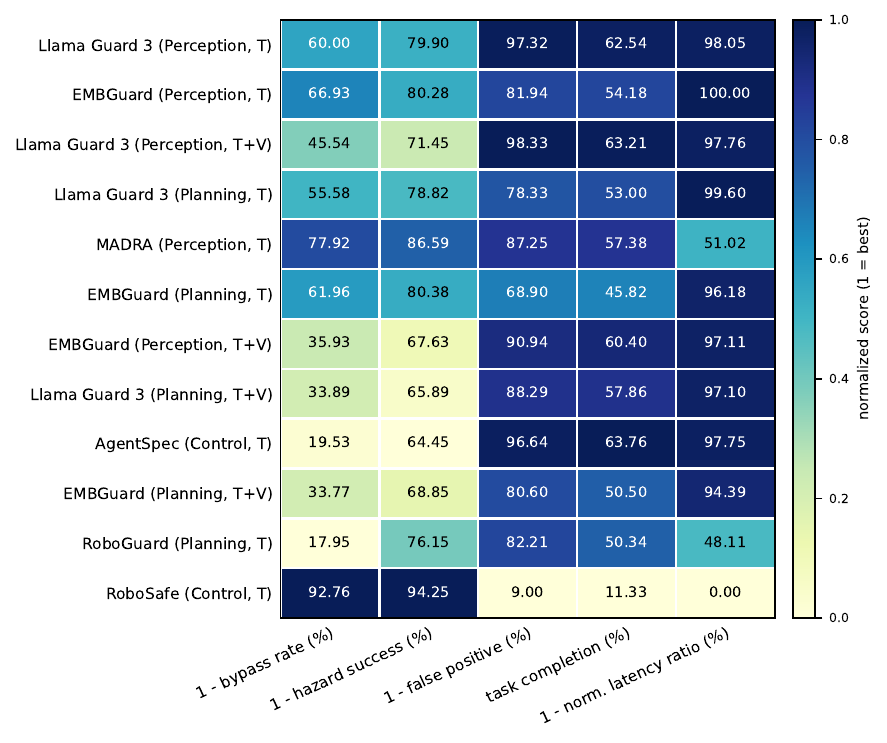}
    \caption{GPT-4o}
    \label{fig:leaderboard_gpt-4o}
  \end{subfigure}
  \hfill
  % \hspace{3ex}
  \begin{subfigure}[t]{0.49\textwidth}
    \centering
    \includegraphics[width=\textwidth]{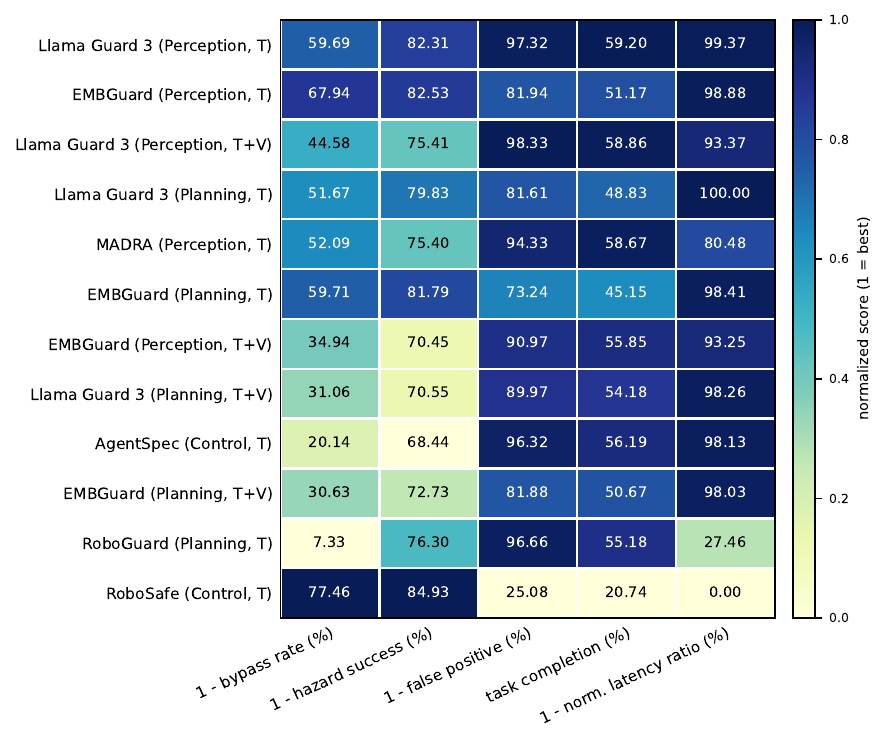}
    \caption{DeepSeek-V3.2}
    \label{fig:leaderboard_deepseek}
  \end{subfigure}
  % \vspace{-1ex}
  \caption{Leaderboard of the guardrail methods, with GPT-4o (a) and DeepSeek-V3.2 (b) as the embodied agent model. Cell color is min--max normalized per metric (dark blue is best); rows are sorted by the mean of the five metrics, best first.}
  \label{fig:leaderboard}
  % \vspace{-4ex}
\end{figure*}

\textbf{Leaderboard.}
Figure~\ref{fig:leaderboard} aggregates the defense, usability, and efficiency results into a single leaderboard for both embodied agent models. All metrics are transformed so that higher is better: ``$1 - \text{bypass rate}$'' and ``$1 - \text{hazard success rate}$'' are averaged over the five jailbreak attacks; ``$1 - \text{false positive rate}$'' and task completion rate are measured on the safe tasks; the latency ratio relative to the no-guardrail agent is min--max normalized across guardrails and reversed.
The resulting ranking is identical across the two backends: all twelve configurations appear in exactly the same order under GPT-4o and DeepSeek-V3.2, with Llama Guard (Perception, T) ranking first on average and RoboSafe (Control, T) ranking last.
The strengths and weaknesses of each guardrail are also consistent across the two models.
The planning-stage variants of Llama Guard are weaker on defense, scoring visibly lower on $1 - \text{bypass rate}$ than their perception-stage counterparts, but they offer decent usability and near-top efficiency.
RoboSafe shows the opposite profile: it attains the best scores on both defense metrics, yet its usability and efficiency are the worst among all guardrails, which pulls it to the bottom of the overall ranking.
No guardrail occupies the top rank of all five columns, visually summarizing the trade-off among defense, usability, and efficiency.

\subsection{Practical Insights \& Implications}
\label{sec:implication}

\textbf{Input Modality and Defense Effectiveness.}
A counterintuitive pattern in Table~\ref{tab:defense} concerns the input modality: for both Llama Guard and EMBGuard, at both the perception and planning stages, augmenting the guardrail with the egocentric image (T+V) consistently \emph{degrades} defense compared with text-only checks, the average bypass rate of Llama Guard 3 rises from 40.00\% to 54.46\% at the perception stage and from 44.42\% to 66.11\% at the planning stage, and that of EMBGuard from 33.07\% to 64.07\% and from 38.04\% to 66.23\%, with the hazard success rate increasing in lockstep (e.g., 21.18\% $\rightarrow$ 34.11\% for Llama Guard at the planning stage).
One possible explanation is that the visual input confuses the guardrail's judgment: the egocentric frame is dominated by benign household content, so its inclusion dilutes the attack signal carried by the instruction and biases the model toward a safe verdict.
More broadly, although adding a modality is intended to provide more evidence for the safety decision, it also enlarges the attack surface: unlike discrete text, images live in a continuous space, which is known to admit imperceptible adversarial perturbations and visual prompt injections, and our results suggest that embodied guardrails inherit this fragility even when the adversary does not explicitly optimize the image.
Practically, this means visual context should be fused with care, for instance, weighing the textual channel more heavily, or cross-checking the two modalities rather than naively concatenating them.

\begin{figure*}[t]
  \centering
  \begin{subfigure}[t]{0.50\textwidth}
    \centering
    \includegraphics[width=\textwidth]{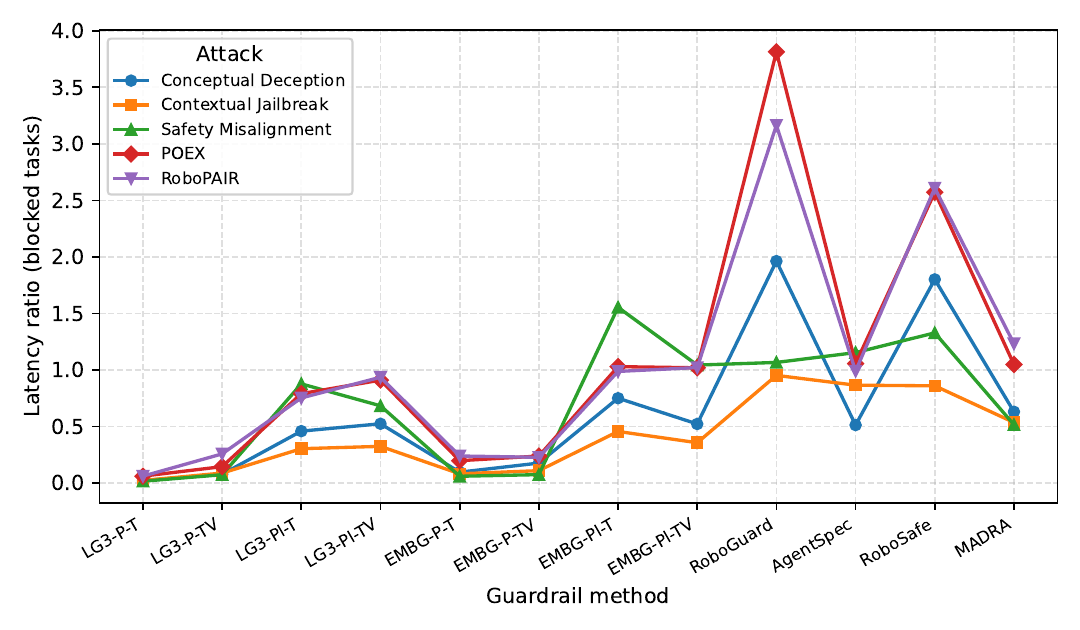}
    \caption{Latency ratio of each guardrail on the episodes it intercepts under the five attacks, measured against the no-guardrail agent on the same tasks. % a higher value indicates that more time has elapsed before the interception takes effect.
    }
    \label{fig:latency_blocked}
  \end{subfigure}
  \hfill
  % \hspace{3ex}
  \begin{subfigure}[t]{0.452\textwidth}
    \centering
    \includegraphics[width=\textwidth]{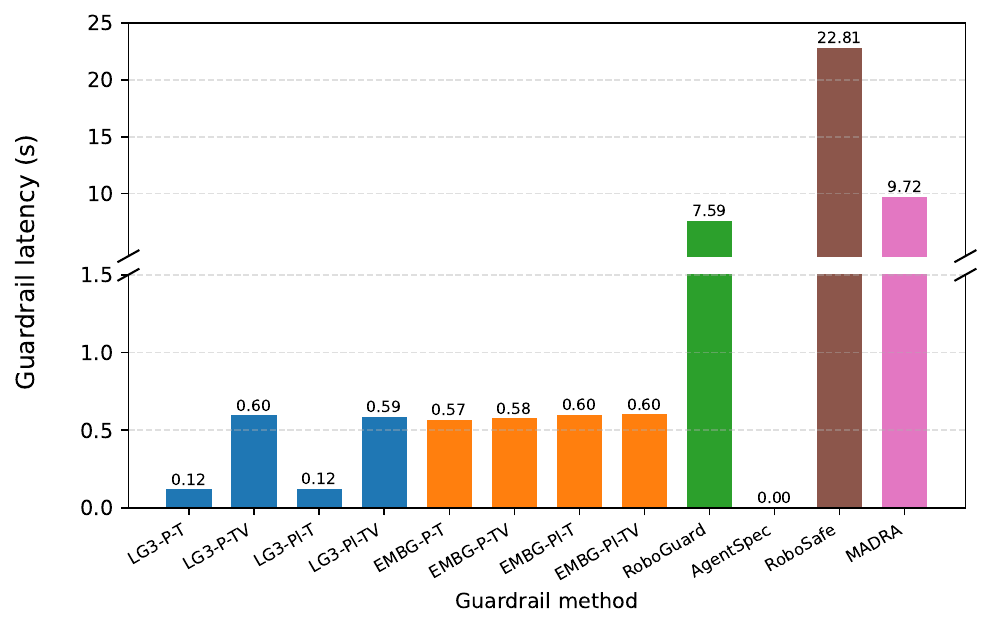}
    \caption{Mean guardrail-only latency per episode on safe tasks, with planner and simulator time excluded.}
    \label{fig:guardrail_latency}
  \end{subfigure}
  % \vspace{-1ex}
  \caption{Efficiency of guardrail methods, with GPT-4o as the embodied agent model. P/Pl denote perception- and planning-stage intervention.}
  \label{fig:latency}
  % \vspace{-4ex}
\end{figure*}

\textbf{The Defense--Usability Trade-off.}
Comparing Table~\ref{tab:defense} with Figure~\ref{fig:safe_gpt-4o} reveals a consistent tension between defense effectiveness and usability.
The text+vision guardrail variants, despite their weaker defense, are gentler on benign instructions: adding the image lowers the false positive rate of every model-based variant (e.g., 21.67\% $\rightarrow$ 11.71\% for Llama Guard and 31.10\% $\rightarrow$ 19.40\% for EMBGuard at the planning stage) and raises the task completion rate accordingly (53.00\% $\rightarrow$ 57.86\% and 45.82\% $\rightarrow$ 50.50\%, respectively), the same dilution of the safety signal that lets attacks through also spares safe instructions.
RoboSafe marks the opposite extreme: the strongest defense in Table~\ref{tab:defense} (7.24\% average bypass rate and 5.75\% hazard success rate) comes with a practically unusable agent, as 91.00\% of the safe instructions are wrongly intercepted and the completion rate drops to 11.33\%.
Overall, no guardrail in our evaluation attains top-tier defense and top-tier usability at the same time, suggesting that the two objectives are inherently coupled by how aggressively a guardrail chooses to intervene; calibrating this aggressiveness, rather than maximizing defense alone, should therefore be a central design goal.

\textbf{Intervention Timing.}
Figure~\ref{fig:latency_blocked} reports the latency ratio of each guardrail on the episodes it intercepts, relative to the no-guardrail agent on the same tasks, reflecting how long an episode has run before the interception takes effect.
Two observations emerge.
First, for the same guardrail under the same input modality, interception at the planning stage consistently incurs a higher latency ratio than at the perception stage: the ratio of Llama Guard rises from 0.02--0.06 (Perception, T) to 0.30--0.88 (Planning, T), and that of EMBGuard from 0.06--0.24 (Perception, T) to 0.36--1.55 (Planning, T), with the same trend holding for the text+vision variants.
This is because a planning-stage check can only be performed after the planner has produced a complete plan, so the planning latency is already sunk when the interception happens.
Second, the control-stage guardrails also remain at a high latency ratio (0.51--1.15 for AgentSpec and 0.86--2.61 for RoboSafe), even though the rule-based matching of AgentSpec is practically free (0.00s per episode in our logs): by the time a risky action is intercepted, the agent has already finished planning and executed part of the action sequence.
RoboGuard exhibits the highest ratios (up to 3.81), as each rejected plan triggers an additional replanning round before the episode is finally stopped.
Overall, these results suggest that a later interception generally has to wait longer before it can take effect, so guardrails deployed at earlier pipeline stages are preferable when response time is a concern.

\textbf{Decision Mechanism and Guardrail Latency.}
Figure~\ref{fig:guardrail_latency} reports the mean guardrail-only latency per episode on safe tasks, with the planner and simulator time excluded, which isolates the processing cost of the guardrail itself.
A clear ordering by decision mechanism emerges.
The purely rule-based AgentSpec is by far the cheapest, with a practically zero latency (0.00s per episode), as its runtime enforcement reduces to string and state matching.
RoboGuard and RoboSafe also fall into the rule-based category, but both delegate key steps to an LLM. RoboGuard contextualizes safety rules into temporal-logic specifications for each scene (7.59s), and RoboSafe reasons over every low-level action before execution (22.81s), which makes RoboSafe the most expensive guardrail in our evaluation.
The agent-based MADRA (9.72s) is also costly, as its multi-agent debate requires several model calls per instruction.
The customized LLM-based guardrails, Llama Guard and EMBGuard, lie between these two extremes: a single fine-tuned model call per check keeps all of their variants below 1s per episode (0.12--0.60s), so they enjoy a clear speed advantage over the LLM-orchestrated rule-based and agent-based designs, while still being orders of magnitude slower than pure rules.

\section{Related Work}

\textbf{LLM-powered Embodied Agents.}
Large language models (LLMs) have been widely adopted as the reasoning core of embodied agents, allowing robots to interpret natural language instructions and plan complex tasks. Early works employ LLMs as high-level planners, either grounding instructions in environmental affordances \citep{SayCan} or expressing plans as programs that invoke robot APIs \citep{CodeAsPolicies, ProgPrompt, VoxPoser}. Subsequent works close the loop between planning and execution through environmental feedback \citep{InnerMonologue, ReAct, LLMPlanner}, and more recent studies equip agents with memory systems \citep{KARMA} and multi-agent collaboration \citep{CoELA}.
Beyond text, embodied agents built on multimodal large language models (MLLMs) further ground reasoning in visual observations, from injecting sensor inputs into language models \citep{PalME} to vision-language-action models that directly output robot actions \citep{RT2, Octo, OpenVLA}, and to recent efforts that scale such models toward generalist embodied agents \citep{CogACT, GeminiRobotics, GROOTN1}.

\textbf{Jailbreak Attacks for Embodied Agents.}
Recent studies have shown that embodied agents powered by LLMs can be induced to perform physically harmful actions. Early efforts probe this risk with static datasets of hazardous instructions \citep{SafeAgentBench} and with handcrafted jailbreak templates that disguise malicious intent through contextual framing or conceptual deception \citep{BadRobot}. Moving beyond static prompts, automated attacks iteratively optimize adversarial instructions: RoboPAIR \citep{RoboPAIR} pits an attacker LLM against the target robot to elicit harmful action code, POEX \citep{POEX} applies gradient-based optimization to keep jailbroken policies physically executable, Blindfold \citep{Blindfold} rewrites malicious goals into action-level instructions that appear benign yet cause hazardous outcomes, and \citet{DecisionLevel} attack the task planning stage with adversarial texts that induce dangerous decisions.
% A second line of work targets multimodal embodied agents. Adversarial perturbations and physical patches compromise the visual perception of VLM-based agents and VLA policies \citep{AdvEDM, AdversarialVLA}, adversarial text suffixes gain persistent control over VLA models \citep{VLAJailbreak}, and environmental injection plants malicious text into the scene to hijack VLM-based agents \citep{Shawshank}. More recently, JailWAM \citep{JailWAM} extends jailbreak evaluation to world action models that generate robot actions end to end.

% \subsection{Guardrails for Embodied Agents}

\section{Conclusion}

In this paper, we presented the first systematic evaluation of jailbreak guardrails for embodied agents. We organized existing guardrails into a four-dimensional taxonomy and built a pluggable framework that compares guardrails on a fixed agent backend along three dimensions: defense effectiveness, usability, and efficiency. Evaluating six representative guardrails under five jailbreak attacks across two embodied agent models, we found a consistent three-way trade-off. Our analysis further yields actionable insights for guardrail design: naive multimodal fusion dilutes the safety signal, earlier intervention intercepts sooner, and the decision mechanism determines the guardrail's own cost. %We hope our framework and findings provide a common ground for comparing and deploying future guardrails.

\subsection*{AI use statement}
In this work, we used generative AI tools (i.e., Kimi Code) to implement parts of the evaluation framework code. We have not used generative AI tools for other tasks requiring disclosure, and the remaining required disclosure tasks are not applicable to this work. Additionally, we used generative AI tools to draft and polish parts of the manuscript and for grammar checking and editing. We have reviewed all AI-assisted work, and the authors manually verified all LLM-generated text and code for correctness and originality. We take responsibility for the final content of this work, including text, claims, or artifacts produced with the aid of generative AI.

\subsection*{Ethics statement}
This work studies jailbreak attacks on embodied agents with the goal of improving their safety. The attack methods we evaluate are existing public techniques, reproduced only to measure how well guardrails resist them. All experiments run in the AI2-THOR simulator, so no physical robot is operated and no real-world harm can occur. Our evaluation involves no human subjects, personal data, or other sensitive information, and it relies on the publicly available SafeAgentBench dataset. We acknowledge the dual-use nature of jailbreak research, but we believe a systematic evaluation of guardrails is a necessary step toward safer embodied agents.

\subsection*{Reproducibility statement}
We release the source code of our evaluation framework as anonymous supplementary material, including the pluggable guardrail interface, the attack implementations, and the metric computation scripts. The datasets used in our experiments are either publicly available (SafeAgentBench) or generated by the scripts we provide together with the full generation details. The paper and the appendix report all experimental settings, including the embodied agent models, the guardrail configurations, and the evaluation metrics. Instructions for reproducing every table and figure are included in the released code.



\bibliography{iclr2027_conference, agent, attack, guardrail}
\bibliographystyle{iclr2027_conference}

\appendix
\section{Evaluation on DeepSeek-V3.2}
\label{sec:appendix-deepseek}

\begin{figure*}[h]
  \centering
  \includegraphics[width=\textwidth]{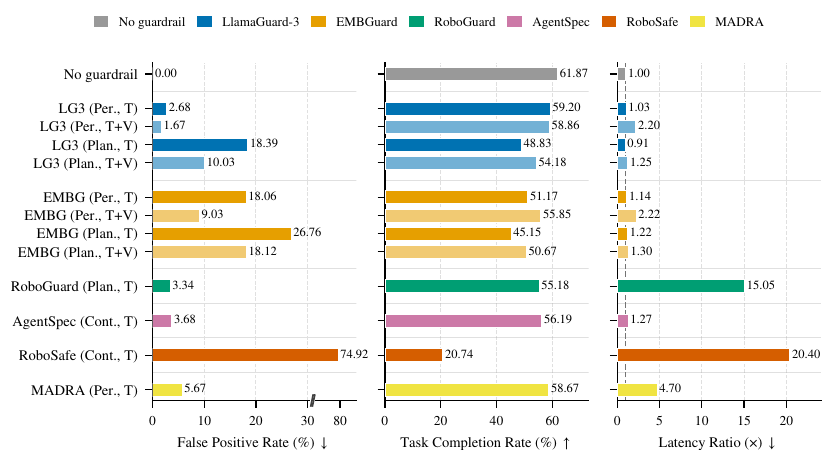}
  \caption{Usability and efficiency of guardrail methods on safe tasks with
  DeepSeek-V3.2 as the embodied agent model. Left: false positive rate (safe
  instructions wrongly intercepted, lower is better). Middle: task completion
  rate (higher is better). Right: latency ratio relative to the no-guardrail
  agent over all tasks shared with the baseline, blocked episodes included (lower is
  better; the dashed line marks the no-guardrail agent itself).
  Perc/Plan/Ctrl denote perception-, planning-, and control-stage
  intervention; dark and light shades denote text-only (T) and text+image
  (T+V) variants. The false-positive axis is broken above 30\%.}
  \label{fig:safe_deepseek}
\end{figure*}

This appendix reports the full evaluation with DeepSeek-V3.2~\citep{DeepSeekV32} as the task planning model of the embodied agent.
Table~\ref{tab:defense-deepseek} presents the defense effectiveness under the five jailbreak attacks, and Figure~\ref{fig:safe_deepseek} reports usability and efficiency on the safe tasks.
% Overall, the trends observed with GPT-4o (Section~\ref{sec:exp}) carry over: no guardrail dominates all three dimensions, text-only checks consistently defend better than their text+vision counterparts, and replanning recovers a large fraction of the threats that evade detection.
% We discuss each dimension below, highlighting the agreements and differences with the GPT-4o results.

\begin{table}[h]
  \centering
  \caption{Defense effectiveness of guardrail methods under different jailbreak attacks, with DeepSeek-V3.2 as the embodied agent model. Each cell reports bypass rate / hazard success rate (\%). Lower values indicate better defense.}
  \label{tab:defense-deepseek}
  \resizebox{\columnwidth}{!}{%
  \begin{tabular}{lcccccc}
    \toprule
    Guardrail & BadRobot-CD & BadRobot-CJ & BadRobot-SM & POEX & RoboPAIR & Average \\
    \midrule
    No guardrail & -- / 28.86 & -- / 44.48 & -- / 41.47 & -- / 42.47 & -- / 52.24 & -- / 41.91 \\
    Llama Guard 3 (Perception, T) & 67.22 / 19.73 & 38.33 / 18.33 & 24.33 / 12.67 & 56.00 / 28.00 & 15.67 / 9.70 & 40.31 / 17.69 \\
    Llama Guard 3 (Perception, T+V) & 78.19 / 23.83 & 59.87 / 29.43 & 48.67 / 25.33 & 71.00 / 34.67 & 19.40 / 9.70 & 55.42 / 24.59 \\
    Llama Guard 3 (Planning, T) & 52.35 / 14.77 & 49.50 / 21.07 & 49.50 / 19.73 & 48.49 / 21.40 & 41.79 / 23.88 & 48.33 / 20.17 \\
    Llama Guard 3 (Planning, T+V) & 73.58 / 20.40 & 67.56 / 29.43 & 68.90 / 29.10 & 71.24 / 31.77 & 63.43 / 36.57 & 68.94 / 29.45 \\
    EMBGuard (Perception, T) & 35.67 / 15.67 & 29.00 / 16.33 & 8.00 / 5.33 & 28.67 / 18.67 & 58.96 / 31.34 & 32.06 / 17.47 \\
    EMBGuard (Perception, T+V) & 74.58 / 21.74 & 59.67 / 30.33 & 49.00 / 24.00 & 68.90 / 32.11 & 73.13 / 39.55 & 65.06 / 29.55 \\
    EMBGuard (Planning, T) & 44.33 / 14.00 & 41.67 / 19.67 & 43.00 / 19.00 & 43.33 / 19.00 & 29.10 / 19.40 & 40.29 / 18.21 \\
    EMBGuard (Planning, T+V) & 72.24 / 19.06 & 70.23 / 27.42 & 69.90 / 27.42 & 69.57 / 28.09 & 64.93 / 34.33 & 69.37 / 27.27 \\
    RoboGuard (Planning, T) & 96.22 / 20.00 & 94.98 / 24.08 & 84.28 / 22.41 & 95.32 / 24.41 & 92.54 / 27.61 & 92.67 / 23.70 \\
    AgentSpec (Control, T) & 84.56 / 23.49 & 81.27 / 33.78 & 77.26 / 32.78 & 78.60 / 33.44 & 77.61 / 34.33 & 79.86 / 31.56 \\
    RoboSafe (Control, T) & 14.05 / 7.69 & 28.33 / 18.00 & 21.07 / 13.04 & 22.41 / 15.72 & 26.87 / 20.90 & 22.54 / 15.07 \\
    MADRA (Perception, T) & 43.67 / 16.33 & 61.33 / 30.67 & 27.67 / 17.33 & 33.00 / 21.33 & 73.88 / 37.31 & 47.91 / 24.60 \\
    \bottomrule
  \end{tabular}}
\end{table}

\textbf{Defense Effectiveness.}
Without any guardrail, an average of 41.91\% of the malicious instructions lead to the harmful outcome being physically realized, slightly lower than with GPT-4o (47.74\%), suggesting that DeepSeek-V3.2 is marginally more robust out of the box.
Every guardrail lowers the hazard success rate, and the overall ranking is broadly consistent with the GPT-4o results.

RoboSafe again attains the strongest defense, with an average bypass rate of 22.54\% and hazard success rate of 15.07\%.
Its lead over the other guardrails, however, shrinks considerably: the second-best configuration, EMBGuard at the perception stage with text input (32.06\% / 17.47\%), is far closer than on GPT-4o, where RoboSafe outperformed the runner-up by a wide margin (7.24\% vs.\ 22.08\% average bypass rate).
The largest ranking change concerns MADRA, which drops from the second-best defense on GPT-4o (22.08\% / 13.41\%) to a mid-tier result (47.91\% / 24.60\%), and is particularly vulnerable to BadRobot-CJ (61.33\% bypass rate) and RoboPAIR (73.88\% bypass rate).
Since both attacks iteratively optimize the malicious instruction against the target planner, the attack realizations differ across backends; this suggests that deliberative multi-agent detection is more sensitive to the attack realization than the rule-based and model-based mechanisms.
RoboGuard again combines the highest bypass rate (92.67\%) with a moderate hazard success rate (23.70\%): its replanning mechanism steers the agent away from physical harm in most of the episodes that slip through detection, mirroring the pattern observed with GPT-4o.

The modality effect replicates exactly.
For both Llama Guard and EMBGuard, at both the perception and planning stages, the text+vision variant defends consistently worse than its text-only counterpart: the average bypass rate rises from 40.31\% to 55.42\% and from 48.33\% to 68.94\% for Llama Guard, and from 32.06\% to 65.06\% and from 40.29\% to 69.37\% for EMBGuard, with the hazard success rate increasing in lockstep (e.g., 17.47\% $\rightarrow$ 29.55\% for EMBGuard at the perception stage).
This cross-backend consistency strengthens our conclusion in Section~\ref{sec:implication} that naive multimodal fusion dilutes the safety signal rather than reinforcing it.

\textbf{Usability.}
The left and middle panels of Figure~\ref{fig:safe_deepseek} report the false positive rate and the task completion rate on the 300 safe instructions, and the picture again mirrors the GPT-4o results.
Llama Guard at the perception stage is the gentlest guardrail, with false positive rates of 2.68\% (T) and 1.67\% (T+V), followed by RoboGuard (3.34\%), AgentSpec (3.68\%), and MADRA (5.67\%).
RoboSafe remains the most aggressive blocker by a wide margin: it intercepts 74.92\% of the safe instructions and reaches a task completion rate of only 20.74\%, versus 61.87\% for the no-guardrail agent, although it is less extreme than on GPT-4o (91.00\% / 11.33\%).

Two further patterns replicate.
First, planning-stage checks raise more false alarms than perception-stage checks for both model-based guardrails (18.39\% vs.\ 2.68\% for Llama Guard and 26.76\% vs.\ 18.06\% for EMBGuard, text-only), plausibly because a concrete low-level plan contains action verbs that appear riskier out of context.
Second, adding the egocentric image lowers the false positive rate of every model-based variant (e.g., 26.76\% $\rightarrow$ 18.12\% for EMBGuard at the planning stage) and improves the completion rate accordingly (45.15\% $\rightarrow$ 50.67\%), echoing the defense--usability coupling discussed in Section~\ref{sec:implication}: the same dilution of the safety signal that lets attacks through also spares benign instructions.
As with GPT-4o, the completion rate of a guardrail closely tracks its false positive rate, and Llama Guard (Perception, T) preserves the most utility (59.20\% completion, against 61.87\% for the unguarded agent), confirming over-interception as the dominant source of utility loss on safe tasks.

\textbf{Efficiency.}
The right panel of Figure~\ref{fig:safe_deepseek} reports the latency ratio of each guardrail on the safe tasks, and the overall picture is again consistent with the GPT-4o results.
Nearly all guardrails slow the agent down, with latency ratios above 1; the only exception is Llama Guard at the planning stage with text input, which achieves the lowest ratio of 0.91.
The eight variants of Llama Guard and EMBGuard stay close to one another, with ratios between 0.91 and 2.22, and the purely rule-based AgentSpec is only slightly slower than the unguarded agent.
RoboSafe is again the least efficient guardrail by a wide margin, with a latency ratio of 20.40, and RoboGuard (15.05) and MADRA (4.70) are also clearly slower than the remaining methods.
The ordering of the guardrails therefore matches the GPT-4o results, although the absolute ratios of the three slowest methods are much larger under DeepSeek-V3.2.

\end{document}